\pdfoutput=1

\documentclass[11pt]{article}

\usepackage[]{acl}
\usepackage{graphicx}

\usepackage{amssymb}
\usepackage{float}
\usepackage{tabularx, booktabs}
\usepackage{colortbl} 
\usepackage{multirow} 
\usepackage{multicol}
\usepackage{times}
\usepackage{latexsym}
\usepackage{graphicx}
\usepackage{listings}

\usepackage{color}
\usepackage{tcolorbox}
\usepackage{soul} 
\usepackage{CJKutf8}
\tcbuselibrary{listingsutf8} 
\usepackage{adjustbox}

\tcbset{width=0.9\textwidth,boxrule=0pt,colback=red,arc=0pt,auto outer arc,left=0pt,right=0pt,boxsep=5pt}

\usepackage{times}
\usepackage{latexsym}
\usepackage[T1]{fontenc}

\usepackage[utf8]{inputenc}

\usepackage{microtype}

\usepackage{inconsolata}

\usepackage{multirow}
\usepackage{amsmath}
\usepackage{cleveref}
\usepackage{tcolorbox}
\usepackage{xcolor,colortbl}
\usepackage{caption}
\usepackage{booktabs,subcaption,amsfonts,dcolumn}
\usepackage{pifont}
\usepackage{ulem} 
\usepackage{tikz}
\usepackage{bbding}
\usepackage{cancel}
\usepackage{graphicx}
\usepackage{hyperref}  
\usepackage{enumitem}
\usepackage{algorithm}
\usepackage{algpseudocode}
\usepackage{xspace}

\usepackage{CJK}
\usepackage{wrapfig,lipsum}

\newtcolorbox{mybox}[1]{colback=gray!10!white,colframe=gray!75!black, boxrule=0.1mm,fonttitle=\bfseries,title=#1,
width=\linewidth}

\definecolor{TT}{HTML}{c7ceea}
\definecolor{VT}{HTML}{F9D0DA}
\definecolor{VV}{HTML}{c0dad1}
\definecolor{correct}{HTML}{007000}
\definecolor{wrong}{HTML}{cc0000}

\newcommand{\grayc}[0]{\cellcolor[rgb]{0.957,0.957,0.957}}

\newcommand{\dataname}{\textit{Chumor}}

\newcommand{\chenv}[1]{\begin{CJK}{UTF8}{gbsn}#1\end{CJK}}
\newcommand{\trchenv}[1]{\begin{CJK}{UTF8}{bsmi}#1\end{CJK}}
\usepackage{color}
\usepackage{mdframed}

\definecolor{lightgray}{rgb}{0.9,0.9,0.9}

\newtcolorbox{example}{
   boxrule=0pt,
   colback=white,
   colframe=white,
   boxsep=5pt,
   left=2pt,
   right=2pt,
   top=2pt,
   bottom=4pt,
   fontupper=\pcr\small,
   width=\linewidth,  
   before skip=0pt,   
   after skip=0pt,    
}

\title{\dataname\ 2.0: Towards Benchmarking Chinese Humor Understanding}

\author{
    Ruiqi He$^1$ \quad 
    Yushu He$^1$ \quad
    Longju Bai$^{1}$ \quad
    {\bf Jiarui Liu$^{2}$ \quad}
    {\bf Zhenjie Sun$^{1}$ \quad}
    {\bf Zenghao Tang$^{3}$ \quad}\\
    {\bf He Wang$^{2}$ \quad}
    {\bf Hanchen Xia$^{3}$ \quad}
    {\bf Rada Mihalcea$^{1}$ \quad}
    {\bf Naihao Deng$^{1\star}$ \quad}\\
    $^{1}$University of Michigan\quad
    $^{2}$Carnegie Mellon University\quad
    $^{3}$Shanghai Jiaotong University \\
    {\tt \{ruiqih, dnaihao\}@umich.edu}
}

\begin{document}
\maketitle
\def\thefootnote{$\star$}\footnotetext{Corresponding author of this work.}
\begin{abstract}
    Existing humor datasets and evaluations predominantly focus on English, leaving limited resources for culturally nuanced humor in non-English languages like Chinese. 
    To address this gap, we construct \dataname, the first Chinese humor explanation dataset that exceeds the size of existing humor datasets. \dataname\ is sourced from Ruo Zhi Ba (RZB, \chenv{弱智吧}), a Chinese Reddit-like platform known for sharing intellectually challenging and culturally specific jokes. 
    We test ten LLMs through direct and chain-of-thought prompting, revealing that \dataname\ poses significant challenges to existing LLMs, with their accuracy slightly above random and far below human.
    In addition, our analysis highlights that human-annotated humor explanations are significantly better than those generated by GPT-4o and ERNIE\textsubscript{4-turbo}.
    We release \dataname~at \url{https://huggingface.co/datasets/dnaihao/Chumor}, our project page is at \url{https://dnaihao.github.io/Chumor-dataset/}, 
    our leaderboard is at \url{https://huggingface.co/spaces/dnaihao/Chumor}, and our codebase is at \url{https://github.com/dnaihao/Chumor-dataset}.
\end{abstract}

\section{Introduction}



Humor is an intrinsic human trait that touches the core of our social and emotional lives, making it a rich field of study across various disciplines \cite{lefcourt2001humor, mihalcea-strapparava-2005-making, gelkopf2011use, hessel-etal-2023-androids}. 
With the advent of Large Language Models (LLMs), researchers have evaluated LLMs' performance on diverse tasks \cite{liu2023evaluating, deng2024tables, wu-etal-2023-hi}
and observed LLMs' extraordinary performance on many \cite{10.1162/tacl_a_00632}.
In contrast, researchers have observed that LLMs still fail to understand humor \cite{ghanadian-etal-2023-chatgpt}.
However, with all these studies on humor, most evaluations remain in English \cite{radev-etal-2016-humor, hasan-etal-2019-ur}.
This presents a significant gap, particularly for non-English languages like Chinese, where culturally nuanced humor understanding is unexamined.

\begin{table*}[!t]
    \small
    \centering
    \resizebox{\linewidth}{!}{
    \begin{tabular}{lll}
        \toprule
        \multicolumn{3}{l}{\grayc \textbf{Cultural}} \\
       Desc.  &  Require knowledge of specific historical, social, or linguistic contexts. & \\
        \multirow{2}{*}{Ex.} & (zh) \chenv{小明在正月接发竟导致舅舅复活。} & \Cref{fig:culture-example}\\
                     & (en) Xiaoming got hair extensions during the first lunar month, which astonishingly brought his uncle back to life. & \\
       \midrule
       \multicolumn{3}{l}{\grayc \textbf{Situational}} \\
       Desc.  & Involve humor derived from specific contexts, irony, or narrative setups. & \\
       \multirow{2}{*}{Ex.} & (zh) \chenv{真可怕，犯罪嫌疑人就在我们之中,被告席上一名法警对另一名法警说。} & \Cref{fig:situational-example}\\
                     & (en) ``Terrifying, the criminal suspect is right between the two of us,'' said one bailiff to another in the defendant's dock. & \\
       \midrule
       \multicolumn{3}{l}{\grayc \textbf{Pun-based}} \\
       Desc.  & Build on linguistic ambiguity and wordplay, require models to identify dual meanings. & \\
       \multirow{2}{*}{Ex.} & (zh) \chenv{你可以在steam上找到GTA，所以水是DNA。} & \Cref{fig:pun-example}\\
                     & (en) 
You can find GTA on Steam, so water is DNA.  & \\
       \midrule
       \multicolumn{3}{l}{\grayc \textbf{Homophonic}} \\
       Desc.  & Rely on phonetic similarities between words or phrases to create humor.  & \\
       \multirow{2}{*}{Ex.} & (zh) \chenv{家里的猪油没了，小明只能把植物油倒快点当猪油用了。} & \Cref{fig:homophonic-example}\\
                     & (en) With the lard gone, Xiaoming had to pour the vegetable oil quickly to use it like lard. & \\
       \midrule
       \multicolumn{3}{l}{\grayc \textbf{Glyph-based}} \\
       Desc.  & Exploit the structural or visual elements of Chinese characters to create humor.  & \\
       \multirow{2}{*}{Ex.} & (zh) \chenv{我把}\trchenv{電車難題}\chenv{简化了，现在是电车难题。} & \Cref{fig:glyph-example} \\
                     & (en) I simplified the trolley problem (in traditional Chinese), now it's the trolley problem (in simplified Chinese). & \\
       \midrule
       \multicolumn{3}{l}{\grayc \textbf{Cross-lingual}} \\
       Desc.  & Involve humor derived from linguistic or phonetic interplay across multiple languages.  & \\
       \multirow{2}{*}{Ex.} & (zh) \chenv{曹操于城楼上问夏侯}\trchenv{惇}\chenv{：“你瞧到了什么。”夏侯}\trchenv{惇}\chenv{说：“瞧到马岱。”} &  \Cref{fig:cross-example}\\
                     & (en) Cao Cao, from atop the city tower, asked Xia Houdun, ``What did you see?'' Xia Houdun replied, ``I saw Ma Dai.'' & \\
       \bottomrule
    \end{tabular}
    }
    \caption{Different types of jokes. Descriptions (Desc.) explain humor mechanisms.
    Examples (Ex.) illustrate each category. 
    The corresponding explanations can be found in the referenced figures from the rightmost column.}
    \label{tab:joke-type}
\end{table*}


In this paper, we try to bridge this gap by constructing \dataname, a funny and challenging Chinese humor understanding dataset sourced from Ruo Zhi Ba (\begin{CJK}{UTF8}{gbsn}RZB, ``弱智吧''\end{CJK} in Chinese), a Chinese version of Reddit platform known for sharing intellectually challenging and culturally specific jokes.
This platform provides a set of unique Chinese jokes that incorporate the subtleties and intricacies of Chinese humor.
\Cref{tab:joke-type} provides examples of the jokes from RZB.
In addition, \citet{bai2024coig} reveal that tuning LLMs on RZB data yields the best performance on Chinese reasoning tasks compared to other data sources, highlighting the significant value of jokes from RZB.

Unlike existing datasets that focus on tasks such as humor detection, punchline identification, or humor generation, \dataname\ addresses the challenge of humor explanation. This involves not just identifying humor but understanding the reasoning behind it, a task that requires both linguistic and cultural knowledge.
Specifically, \dataname\ tasks the LLMs with determining whether an explanation fully explains the joke.
We source the explanations from GPT-4o and ERNIE\textsubscript{4-turbo}, and have the entire dataset manually annotated by five native Chinese speakers.
We evaluate ten LLMs from various model families, and reveal that all models perform poorly, lagging significantly behind humans on \dataname.
We observe that chain-of-thought prompting does not necessarily improve models performance and can sometimes confuse their reasoning process.
In addition, we conduct a case study in which one of the authors annotates the entire dataset, followed by A/B testing conducted by six native Chinese speakers to compare explanations from GPT-4o versus human, and ERNIE\textsubscript{4-turbo} versus human.
Our results indicate that human-annotated joke explanations are significantly better than those produced by GPT-4o or ERNIE\textsubscript{4-turbo} (\Cref{fig:gpt-4o-preference-eval}), with LLMs yielding winning rates of only 6.2\% for GPT-4o and 5.3\% for ERNIE\textsubscript{4-turbo} compared to humans.


In summary, our contributions are three folds:

\begin{enumerate}[leftmargin=\parindent,align=left,labelwidth=\parindent,labelsep=0pt]
    \item We construct \dataname, a funny and challenging Chinese humor understanding dataset, which is the largest Chinese humor explanation dataset.
    \item We evaluate ten LLMs on \dataname\ and reveal the significant challenges \dataname\ possesses. We highlight that the best accuracy achieved by LLMs is 60.3\%, significantly lower than human's score of 78.3\%.
    \item We demonstrate that chain-of-thought prompting can hurt LLM's performance in humor reasoning, and that human-annotated joke explanations are significantly better than those produced by GPT-4o and ERNIE\textsubscript{4-turbo}urging., encouraging future research on culturally specific humor understanding
\end{enumerate}

\section{Related Works}

\paragraph{Humor Datasets.} Humor analysis in natural language processing (NLP) encompasses a wide range of tasks, each focused on different aspects of humor.
For instance, researchers have proposed datasets such as ``16000 One-Liners'' \cite{mihalcea-strapparava-2005-making}, ``Pun of the Day'' \cite{yang-etal-2015-humor}, and ``Ted Laughter'' \cite{chen-lee-2017-predicting} focused on humor detection to determine whether a given text is humorous or not.
Datasets such as ``Big Bang Theory'' \cite{bertero-fung-2016-deep} aim at pinpointing the punchline in a joke.
Tasks for assessing humor intensity include humor level rating, comparison, and ranking. For example, datasets like HumorNorm \cite{engelthaler2018humor} and \#HashtagWars \cite{potash-etal-2017-semeval} quantify humor scores and compare comedic elements, while UR-FUNNY ranks punchlines based on their perceived impact.
Datasets such as ``Humicroedit'' \cite{hossain-etal-2019-president}, ``$C^{3}$'' \cite{Wang2022CanLM}, and ``TalkFunny'' \cite{Chen_Yuan_Liu_Liu_Guan_Guo_Peng_Liu_Li_Xiao_2024} focus on humor generation, the task of generating or rewriting humorous texts.
In addition, we present a comprehensive overview of the existing datasets related to humor in \Cref{tab:dataset_comparison}.
We highlight that most existing datasets are in English.
Chinese humor, on the other hand, is less explored.
Our dataset, \dataname\, is the first humor explanation dataset in Chinese.

\begin{table}[t]
  \small
  \centering
  \resizebox{\linewidth}{!}{%
    \begin{tabular}{llcrl}
      \toprule
      Dataset & Sources & Lan. & \#(k) & Tasks \\
      \midrule
      One Liners (\citeyear{mihalcea-strapparava-2005-making}) & Web & en & 16 & HR  \\
      \grayc Pun of the Day (\citeyear{yang-etal-2015-humor}) & \grayc Web & \grayc en & \grayc 4.8 & \grayc \begin{tabular}[c]{@{}c@{}}HR\\PD\end{tabular} \\
Big Bang Theory (\citeyear{bertero-fung-2016-deep}) & TV & en & 44 & PD\\
\grayc Ted Laughter (\citeyear{chen-lee-2017-predicting}) & \grayc TED & \grayc en & \grayc 9.4 & \grayc \begin{tabular}[c]{@{}c@{}}HR\\PD\end{tabular}\\
\#HashtagWars (\citeyear{potash-etal-2017-semeval}) & TV & en & 13 & HC \\
\grayc HumorNorm (\citeyear{engelthaler2018humor}) & \grayc CS$^\dagger$ & \grayc en & \grayc 5 & \grayc HC \\
UR-FUNNY (\citeyear{hasan-etal-2019-ur}) & TED & en & 17 & PD \\
\grayc Humicroedit (\citeyear{hossain-etal-2019-president}) & \grayc Reddit & \grayc en & \grayc 15 & \grayc HG \\
rJokes (\citeyear{weller-seppi-2020-rjokes}) & Reddit & en & 57 & HC \\
\grayc Memotion (\citeyear{sharma-etal-2020-semeval}) & \grayc Memes & \grayc en & \grayc 9.8 & \grayc HC \\
MUMOR (\citeyear{10.1007/978-3-030-88480-2_49}) & TV & \begin{tabular}[c]{@{}c@{}}en\\zh\end{tabular} & 30 & HR \\
\grayc NYT-Captions (\citeyear{hessel-etal-2023-androids}) & \grayc NYT & \grayc en & \grayc \begin{tabular}[c]{@{}c@{}}{0.7}\\2.6\end{tabular} & \grayc \begin{tabular}[c]{@{}c@{}}{\bf HE}\\ HC\end{tabular} \\
\midrule
$C^{3}$ (\citeyear{Wang2022CanLM}) & Books & zh & 9.3 & HG \\
\grayc TalkFunny (\citeyear{Chen_Yuan_Liu_Liu_Guan_Guo_Peng_Liu_Li_Xiao_2024}) & \grayc Apps & \grayc zh & \grayc 4.1 & \grayc HG \\
TCHD (\citeyear{10.1145/3539597.3570431}) & -- & zh & 26 & \begin{tabular}[c]{@{}c@{}}HR\\HC\\PD\end{tabular} \\
\grayc TTWS (\citeyear{Zhang2019TellingTW}) & \grayc Books & \grayc zh & \grayc 9.1 & \grayc PD \\
CHM (\citeyear{tseng-etal-2020-development}) & \begin{tabular}[l]{@{}l@{}}Apps\\Web\end{tabular}  & zh & 3.3 & HC \\
\grayc Memeplate (\citeyear{10.1007/978-3-031-17120-8_41}) & \grayc \begin{tabular}[l]{@{}l@{}}Apps\\Web\end{tabular} & \grayc zh & \grayc 5.2 & \grayc HC \\
\dataname~(us) & Web & zh  & 3.3 & {\bf HE}\\

      \bottomrule
    \end{tabular}
  }
  \caption{Existing datasets related to humor. For the shorthands in the table, abbreviations represent the following tasks, 
  HR: humor recognition; 
  PD: punchline detection; HC: humor comparison; HG: humor generation; HE: humor explanation
  $\dagger$: Crowd-source.
  }
\label{tab:dataset_comparison}
\end{table}

\begin{table*}[t]
    \centering
    \small
    \renewcommand{\arraystretch}{1.3}
    \setlength\tabcolsep{3pt}
    \resizebox{\linewidth}{!}{
    \begin{tabular}{p{0.25\linewidth} p{0.70\linewidth} cc}
        \toprule
        \textbf{Joke} & \textbf{Explanation} & \textbf{Label} & \textbf{Source} \\ \midrule

        (zh) \chenv{“你自己掰了一千吨玉米?!”\space“哈哈我瞎掰的”} & 
        (zh) \chenv{这个笑话的幽默之处在于对话中的双关语和出人意料的回答。“掰了一千吨玉米”看似是一个惊人的壮举，而“哈哈我瞎掰的”则瞬间揭露了这其实是一个夸张的玩笑，这种巧妙的反转和语言的戏谑带来了幽默感。} & 
        \multirow{6}{*}{good} & \multirow{6}{*}{E} \\
        \grayc (en) ``You harvested a thousand tons of corn all by yourself?!''
"Haha, I just made it up." & 
        \grayc (en) The humor in this joke lies in the pun and the unexpected response in the conversation. ``Harvested a thousand tons of corn'' seems like an extraordinary achievement, but the reply, ``Haha, I just made it up,'' instantly reveals it to be an exaggerated joke. This clever twist and playful use of language create the comedic effect. & 
         & \\  
        \midrule

        (zh) \chenv{小明突然出现在我面前以至于我突然出现在小明面前。} & 
        (zh) \chenv{这个笑话的幽默之处在于它利用了语言的对称性和意外性。小明突然出现导致叙述者的惊讶反应，而叙述者的惊讶反应又反过来让小明感到意外，\textcolor{red}{\underline{形成了一个有趣的循环}}。} & 
        \multirow{6}{*}{\textcolor{red}{\underline{bad}}} & \multirow{6}{*}{G} \\
        \grayc (en) Xiaoming suddenly appeared in front of me, causing me to suddenly appear in front of him. & 
        \grayc (en) The humor in this joke lies in its use of linguistic symmetry and unexpectedness. Xiao Ming's sudden appearance triggers a surprised reaction from the narrator, which in turn surprises Xiao Ming, \textcolor{red}{\underline{creating an amusing loop}}.& 
         & \\
        \bottomrule
    \end{tabular}
    }
    \\
    \vspace{1em}
    \caption{Examples from \dataname.
    For sources, ``G'' and ``E'' indicate the explanation comes from GPT-4o and ERNIE\textsubscript{4-turbo}, respectively.
    The second example's explanation is bad because the joke does not ``creating an amusing loop''. 
     Instead, it relies on linguistic symmetry and the use of a straightforward fact to subvert expectations. 
     The audience anticipates an unexpected outcome due to the setup, but the latter part ``suddenly appear in front of him'' flips the perspective by stating the straightforward fact that because Xiao Ming is in front of the person so the person is in front of Xiao Ming too.
    }

    \label{table:dataset-example}
\end{table*}

\paragraph{Culturally Specific Datasets.}
Recent works underscore the challenges of culturally specific reasoning in LLMs \citep{shen-etal-2024-understanding, alkhamissi-etal-2024-investigating, pawar2024survey, vayani2024all}. 
These challenges stem from the overrepresentation of Western-centric knowledge and translation artifacts, which limit the fairness and effectiveness of multilingual evaluations \citep{mihalcea2024ai}.
Researchers have proposed various culturally specific datasets such as Global-MMLU \citep{singh2024globalmmluunderstandingaddressing} to evaluate LLMs' cultural knowledge.
\dataname\ adds to this line of effort as it involves rich knowledge specific to Chinese culture.

\section{Chumor Dataset}
\paragraph{Data Collection.}
We construct our dataset by including RZB jokes from ``Best Annual Threads'' between 2018 and 2021 that have been previously crawled\footnote{https://github.com/Leymore/ruozhiba}.
In addition, we directly collect all threads in the ``Moderator's Recommendation'' section from RZB.
Each thread in RZB consists of \begin{CJK}{UTF8}{gbsn}``标题'' (title)\end{CJK}, \begin{CJK}{UTF8}{gbsn}``一楼'' (content)\end{CJK}, and several \begin{CJK}{UTF8}{gbsn}``跟帖'' (follow-up posts)\end{CJK}. 
For threads from Best Annual Threads, the jokes are listed in the follow-up posts, which are selected by the forum moderator.
For threads from Moderator's Recommendation, the jokes consist of the title and the content of each thread.
We remove the content if it repeats the title.

\paragraph{Data Cleaning.}
We store both the title and the content of the raw data.
However, due to the posting restrictions of the platform requiring non-empty content, many posts contain meaningless placeholder texts such as ``.'', ``!'', ``0'', ``RT'', and others. 
We automatically identify and remove these patterns, and only keep the title which is the joke itself.
Due to the length limitations on the original platform, many post titles are truncated from the beginning parts of the content. 
We identify these instances and replace the truncated title with the complete content to get the joke.
We also remove duplicates that appear both in the 
``Moderator's Recommendation'' and the ``Best Annual Posts''.

We manually remove the threads related to forum management and rules, threads that include excessively offensive content, threads with incomplete content, and threads that focus more on philosophical insight rather than humor.
\paragraph{Humor Explanation Classification.}

We design a humor explanation classification task that can be easily used to test LLMs' capabilities in humor understanding.
Specifically, we use two LLMs, GPT-4o and ERNIE\textsubscript{4-turbo} to generae explanations for our collected jokes.
We manually annotate the generated explanations as either ``fully explain the joke'' (good) or ``partially explain or not explain the joke'' (bad) based on a majority vote among five of the authors who are native Chinese speakers.
Each joke, along with its explanation, forms an individual instance in \dataname, leading to a total of 3,339 instances. 
Among these, 1,454 items are labeled as good and 1,887 as bad explanations. 

\paragraph{Data Examples from \dataname .}
We present examples from {\dataname} in \Cref{table:dataset-example}.

\paragraph{Humor Categorization in \dataname.}
We categorize the jokes in RZB into six types in \Cref{tab:joke-type}, with an example provided for each type.

\section{Experiments}
\paragraph{Models.}
We test ten LLMs, five from the open-source LLM families and five from the closed-source LLM families, all capable of handling Chinese.
Specifically, we include the open-source LLMs of \textbf{Yi\textsubscript{34B}} \cite{huggingface2024yi34b} from 01.AI, \textbf{Nemotron\textsubscript{70B}} \cite{huggingface2024llama31nemotron} from NVIDIA, \textbf{Athene\textsubscript{70B}} \cite{huggingface2024athene70b} from Nexusflow, \textbf{Qwen2.5\textsubscript{72B}} \cite{huggingface2024qwen25} from Alibaba, \textbf{Mistral\textsubscript{123B}} \cite{huggingface2024mistral} from Mistral AI, alongside the closed-source LLMs of \textbf{Gemini\textsubscript{1.5-pro}} \cite{google2024gemini15pro} from Google, \textbf{GLM-4\textsubscript{plus}} \cite{glm2024} from Tsinghua University, \textbf{GPT-4\textsubscript{turbo}, GPT-4o} \cite{openai2023gpt4,openai2024gpt4o} from OpenAI, \textbf{ERNIE\textsubscript{4-turbo}} \cite{baidu2024wenxin} from Baidu.
For all the open-source LLMs, we use the instruction-tuned version in our evaluation. 

\paragraph{Evaluation Methods.}

We evaluate these LLMs using two prompting methods: direct prompting (DP) by
\begin{mybox}{Direct Prompting (DP)}
\chenv{
    你将看到一个笑话以及对这个笑话的解释。请判断这个解释是否完全解释了笑话。
    根据判断，选择“完全解释”或“部分/没有解释”，不需要解释为什么对或者不对。

    笑话：\{joke\} \\
    笑话解释：\{explanation\}
}
\end{mybox}

\begin{mybox}{Prompt Translation}
You will see a joke and an explanation of the joke. Please determine whether this explanation fully explains the joke. Based on your judgment, choose either ``fully explain'' or ``partially/does not explain.'' You do not need to explain why it is correct or incorrect.

Joke: \{joke\} \\
Explanation: \{explanation\}
\end{mybox}

\noindent and chain-of-thought (CoT) prompting \citep{wei2022chain} by adding the phrase \chenv{
    ``请逐步思考，写下过程''
} (Please think step by step, write down your reasoning process) before determining the label. 
\Cref{app-sec: prompt} provides the complete prompts.
We calculate accuracy scores as part of our evaluation.
In addition, we provide the false positive rate (FPR), false negative rate (FNR), and Matthews Correlation Coefficient (MCC) in \Cref{app-sec: detailed results} in \Cref{table:performance_categorization}.
The MCC score considers true positives, true negatives, false positives, and false negatives, providing a score between -1 and +1. A score of +1 indicates perfect predictions, 0 reflects random guessing, and -1 means complete disagreement. 
The best MCC score achieved by LLMs is 0.29, which is close to random guessing, and is significantly lower than the human average of 0.60.


\section{Results and Discussions}
\paragraph{Overall Model Performance.}

\begin{figure}[!ht]
    \includegraphics[width=\linewidth]{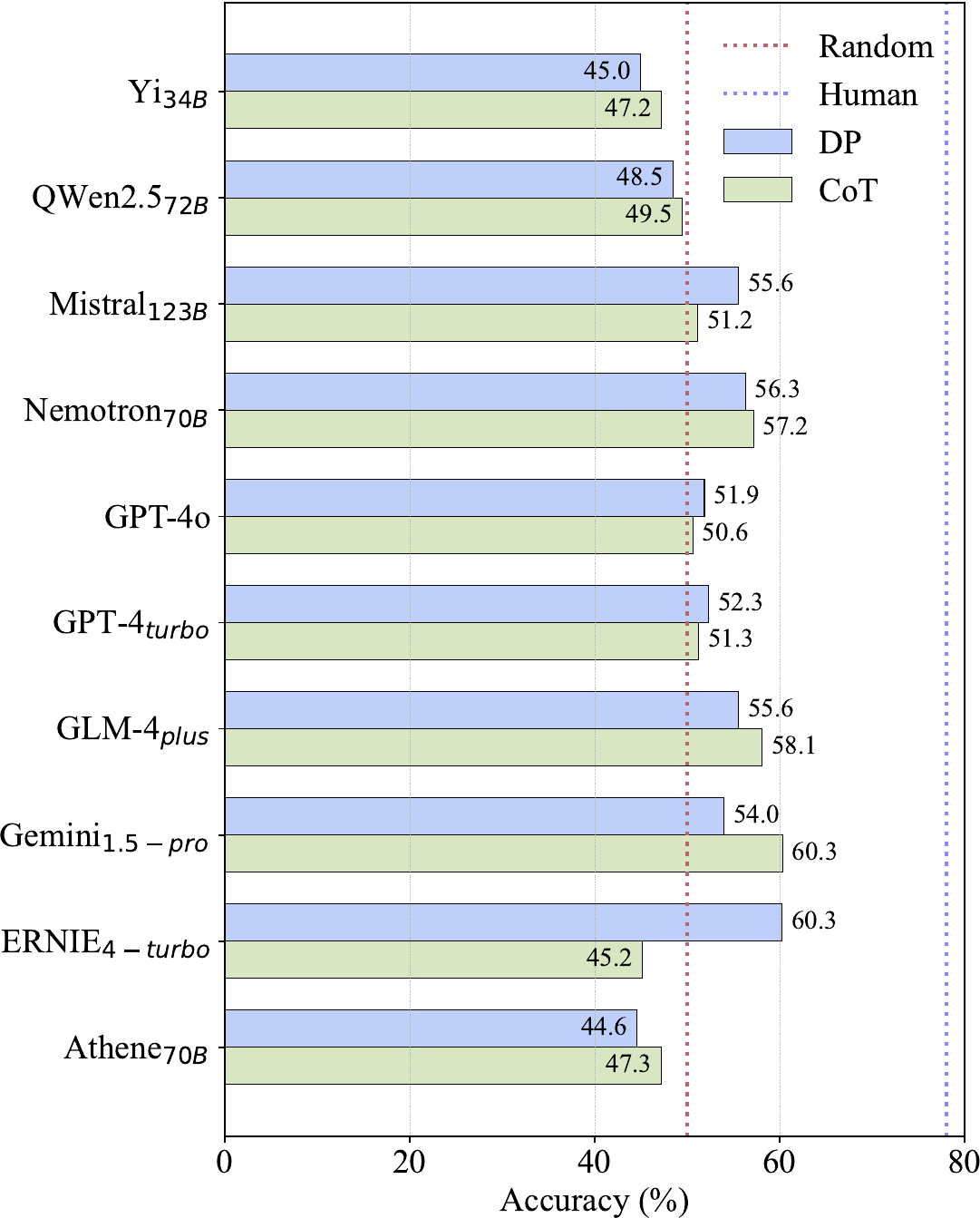}
    \caption{The accuracy of different models' test results in the DP and CoT settings. ERNIE\textsubscript{4-turbo} and Gemini\textsubscript{1.5-pro} achieve the highest accuracy of 60.3\%.}
    \label{fig: accuracy of different models}
\end{figure}

\Cref{fig: accuracy of different models} presents the accuracy of different LLMs on \dataname\ in DP and CoT settings. \Cref{app-sec: detailed results} presents additional results and analysis.

Overall, we observe that all models perform poorly on Chinese humor comprehension, with accuracy scores ranging between 44.6\% and 60.3\%. 
ERNIE\textsubscript{4-turbo} and Gemini\textsubscript{1.5-pro} achieve the highest accuracy of 60.3\%, and are just 10 points above the random baseline and far below human performance of 78.3\%, highlighting the difficulty of \dataname\ and the limitations of these LLMs in understanding Chinese humor.

\subsection{Error analysis by joke type}



\begin{figure}[!t]
    \includegraphics[width=\linewidth]{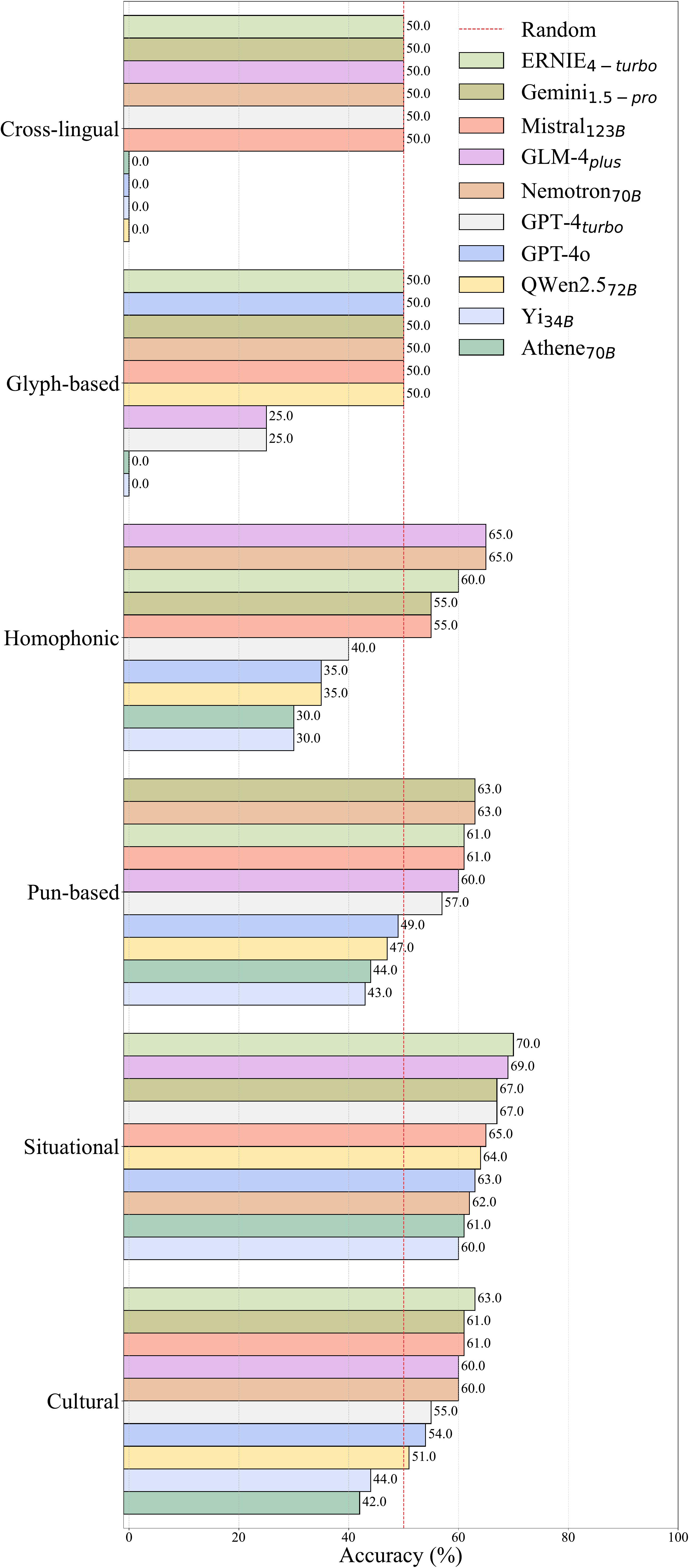}
    \caption{DP accuracy on different joke types (\%). We highlight that model performance varies significantly across different joke types.}
    \label{fig:dp}
\end{figure}

To better understand how LLMs perform on each joke type listed in \Cref{tab:joke-type}, we sample 200 jokes for error analysis.
\Cref{fig:dp} and \Cref{fig:cot} in \Cref{app-sec: detailed results} present the results.

We highlight that model performance varies significantly across different joke types. While models generally perform well on \textit{Situational} jokes, achieving 60.0\% to 70.0\% accuracy in both DP and CoT settings, their performance difference on other joke types is more pronounced. For instance, GLM-4\textsubscript{plus} achieves 65.0\% accuracy on \textit{Homophonic} jokes in the DP setting, whereas Yi\textsubscript{34B} only reaches 30.0\%. Nemotron\textsubscript{70B} performs well on \textit{Cultural} jokes in the CoT setting with 72.0\% accuracy, but Athene\textsubscript{70B} and ERNIE\textsubscript{4-turbo} achieve with only 43.0\% and 42.0\%, respectively.
Such performance variance highlights LLMs' varied capabilities in specific domains such as cultural reasoning and situational reasoning, revealing the respective limitations of these LLMs.


\subsection{Have LLMs achieved human-level understanding of humor?}

\paragraph{Answer: No.}
To compare the performance of LLMs with humans, we conduct a human study involving three Chinese native speakers unfamiliar with this work to annotate a randomly chosen subset of 200 examples. 
Human annotators demonstrate significantly better performance, with an average accuracy of 78.3\% and an MCC socre of 0.60, significantly better than the LLMs' best performance of 60.3\% accuracy and an MCC score of 0.29 (\Cref{fig: MCC of different models} in \Cref{app-sec: detailed results}). Our results indicate that there is a large room of performance improvement for LLMs on Chinese humor understanding.

\subsection{Does chain-of-thought (CoT) help LLMs' humor understanding?}

\paragraph{Answer: No.}

\begin{figure*}[t]
    \includegraphics[width=\linewidth]{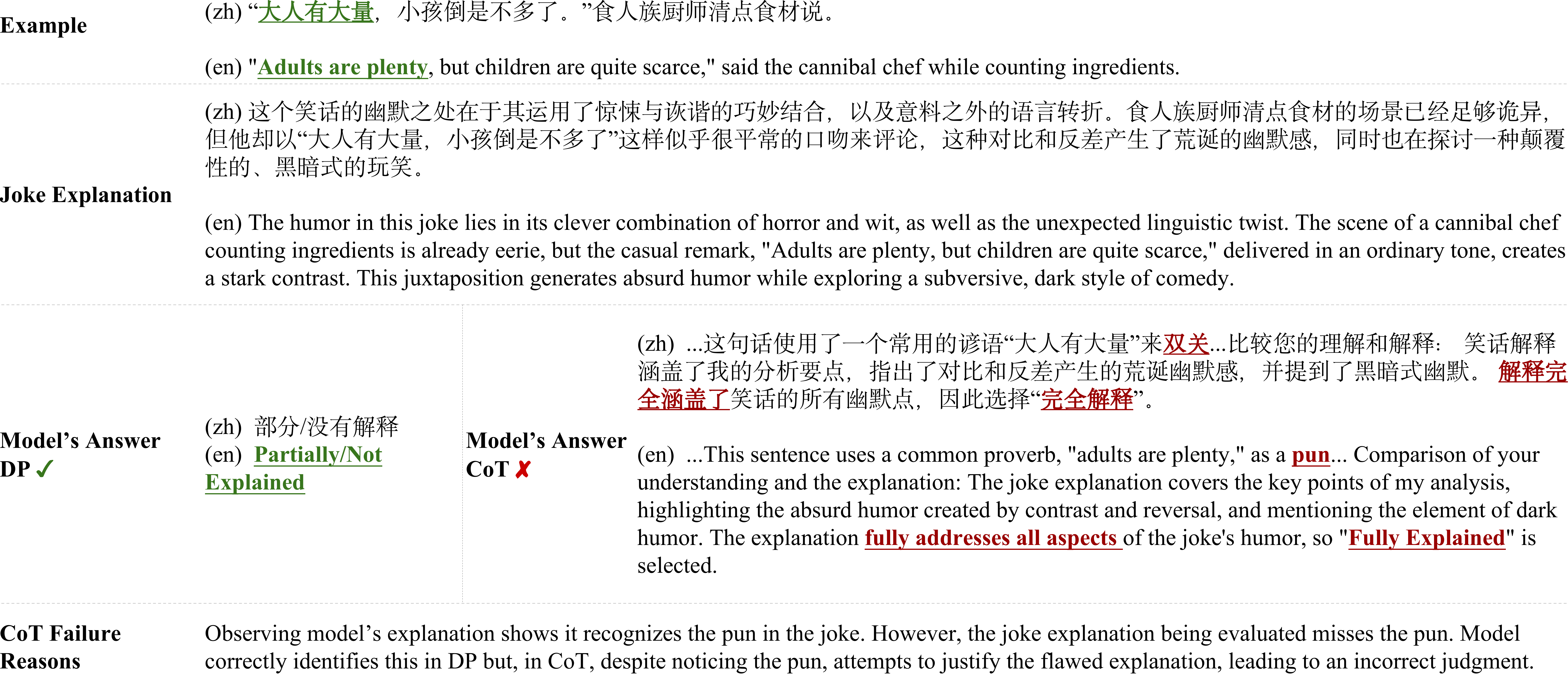}
    \caption{Over-analyzing example by GPT-4o.
    The GPT-4o model chooses the correct answer in the DP prompting, but chooses the incorrect answer due to over-analyzing in the CoT prompting.}
    \label{fig:over_explain_ex}
    \vspace{-0.5em}
\end{figure*}

We observe that CoT does not necessarily improve model performance and, in most cases, even leads to performance decay. 
For instance, as shown in \Cref{fig: accuracy of different models}, the accuracy of ERNIE\textsubscript{4-turbo} decreases from 60.3\% to 45.2\% when we switch to CoT prompting,
Mistral\textsubscript{123B}'s performance drops from 55.6\% to 51.2\%, GPT-4o's performance drops from 51.9\% to 50.6\%, GPT-4\textsubscript{turbo}'s performance falls from 52.3\% to 51.3\%.
Moreover, the MCC scores present a clearer trend of performance decline under CoT prompting. As shown in \Cref{fig: MCC of different models} in \Cref{app-sec: detailed results}, eight of the ten LLMs' MCC scores decrease under CoT prompting. We hypothesize that CoT prompts may not help the model's reasoning when the model lacks a fundamental grasp of humor understanding.

We observe that under CoT prompting, models like GPT-4o tend to justify incorrect explanations as ``correct'', leading to an increase in false-positive rate from 80.0\% for DP prompting to 85.0\% for CoT prompting (\Cref{table:performance_categorization} in \Cref{app-sec: detailed results}). ERNIE\textsubscript{4-turbo} exhibits the largest false-positive rate, rising from 59.8\% to 96.9\% (\Cref{table:performance_categorization} in \Cref{app-sec: detailed results}).
\Cref{fig:over_explain_ex} provides an example where CoT confuses the GPT-4o model.
Under the DP prompting, the GPT-4o model chooses the answer correctly.
However, CoT prompting causes the model to over-analyze and justify an incorrect explanation.

On the other hand, models like Nemotron\textsubscript{70B} may be overly critical of explanations under CoT prompting, resulting in a false-negative rate from 20.9\% for DP prompting to 46.1\% for CoT prompting (\Cref{table:performance_categorization} in \Cref{app-sec: detailed results}). We highlight that a recent work demonstrates that CoT can degrade performance in tasks requiring subtle comprehension \cite{sprague2024cotcotchainofthoughthelps}, which aligns with our findings on its limitations in humor interpretation.
\Cref{fig:over_critical_ex} in \Cref{app-sec:more-error-cases} discusses an example corresponding to the model being overly critical.

\subsection{Case study: can GPT-4o and ERNIE\texorpdfstring{\textsubscript{4-turbo}}{4-turbo} explain jokes as well as humans?}
\paragraph{Answer: No.}
Apart from testing multiple LLMs on \dataname, we conduct case studies on GPT-4o and ERNIE\textsubscript{4-turbo} to assess the quality of their joke explanations compared to humans. 
We prompt them to explain the humor in two sentences, consistent with the format of human explanations.
Here is the prompt we feed to both LLMs: 
\begin{mybox}{Prompt}
\chenv{请用两句话解释这个笑话的幽默之处: [Joke]}
\end{mybox} 
\noindent, which translates to the following prompt.
\begin{mybox}{Prompt Translation}
Please explain the joke in two sentences: [Joke]
\end{mybox}

\paragraph{Data Annotation.}
As demonstrated by \citet{hessel-etal-2023-androids}, crowd-sourcing typically cannot produce high-quality explanations, following \citet{hessel-etal-2023-androids}, one of the authors annotates all the explanations to ensure the quality and consistency.

This is a substantial effort: the author ended up annotating the explanations for 1,951 jokes. The resulting corpus has a mean of 78 Chinese characters of explanation per joke, and the total length, 151,730 Chinese characters, is comparable in length to a novella\footnote{The total length of our explanations surpasses the Chinese version of {\it The Great Gatsby} (100k Chinese characters), and is about half the length of the Chinese version of {\it Wuthering Heights} (325k Chinese characters).}.


\paragraph{Evaluation Setup.}
To fairly evaluate which explanation is better, we conduct A/B testing by presenting the humor explanation from one LLM and from human to six college students, asking them to annotate their preference of the explanation for each joke.
These college students are native Chinese speakers who grew up in China, therefore they have a deep understanding of the cultural terms and trending terms in China.
We note that the preference annotation requires a substantial effort as each annotator reads through a total length of around 300k Chinese characters\footnote{This is about the same length of the Chinese version of {\it Wuthering Heights} (325k Chinese characters).}.
We end up with three preference annotations for each joke.
The preference annotation achieve a 61.4\% agreement rate among annotators (\Cref{app-sec: agreement-rate}).

We use the winning rate as our measure to compare LLMs' explanation versus human explanation, taking the majority vote among all annotators for each example.
In addition, if all annotators disagree, we assign an ``Undecided'' label.
\Cref{app-sec: annotation-instruction} provides the annotation instructions we present to the annotators.


\begin{figure}[t]
    \centering
    \includegraphics[width=\linewidth]{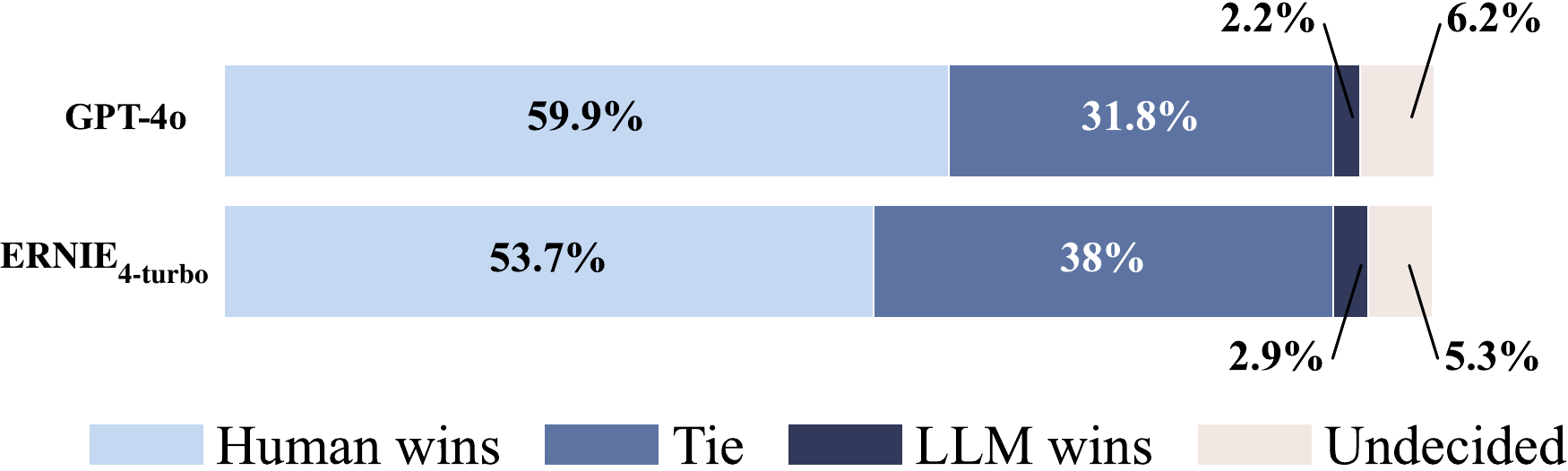}
    \caption{Annotated preference for whether human explanation is preferred (``Human wins'') or the explanation from LLMs is preffered (``LLM wins''). Humans' explanation is significantly preferred over LLMs'.
    }
    \label{fig:gpt-4o-preference-eval}
    \vspace{-0.5em}
\end{figure}

\paragraph{Overall Results.}

\Cref{fig:gpt-4o-preference-eval} reports the wining rate of explanations from human versus GPT-4o and ERNIE\textsubscript{4-turbo}.
We can see that human explanations are significantly better than those from both LLMs, with humans winning over 50\% of the time, while LLMs win in only 2-3\% of cases.



\paragraph{Error Analysis.}


\begin{figure}[t]
    \centering
    \includegraphics[width=0.90\linewidth]{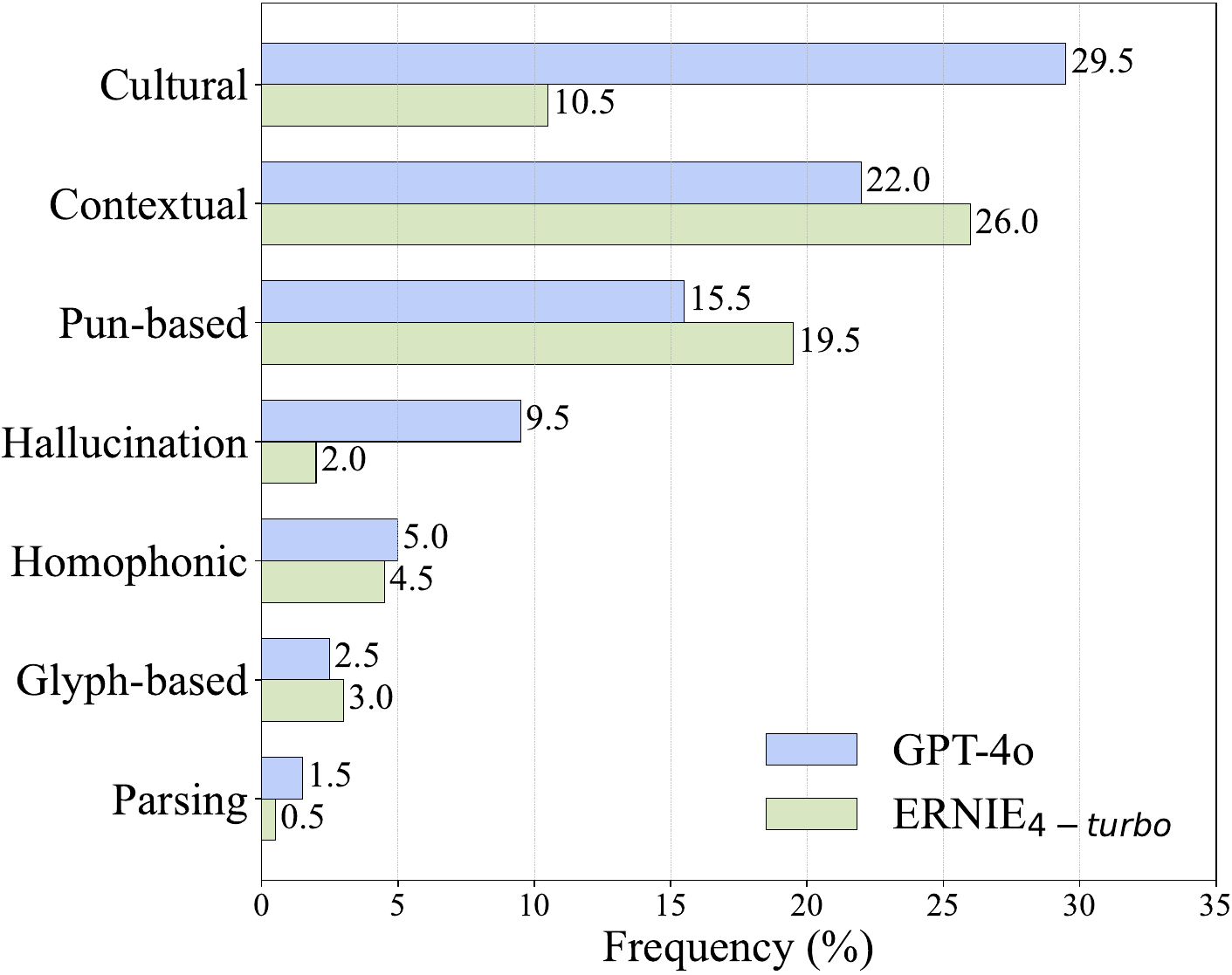}
    \caption{Distribution of error types for GPT-4o and ERNIE\textsubscript{4-turbo}. 
    We sample 200 examples to calculate the distribution of these error types. 
    We note that each example may correspond to multiple error types. We highlight that ERNIE\textsubscript{4-turbo} demonstrates a lower error rate on cultural jokes, while GPT-4o demonstrates a lower error rate on contextual or pun-based jokes.
    }
    \label{fig:distribution}
    \vspace{-0.5em}
\end{figure}

\begin{figure}[t]
    \includegraphics[width=\linewidth]{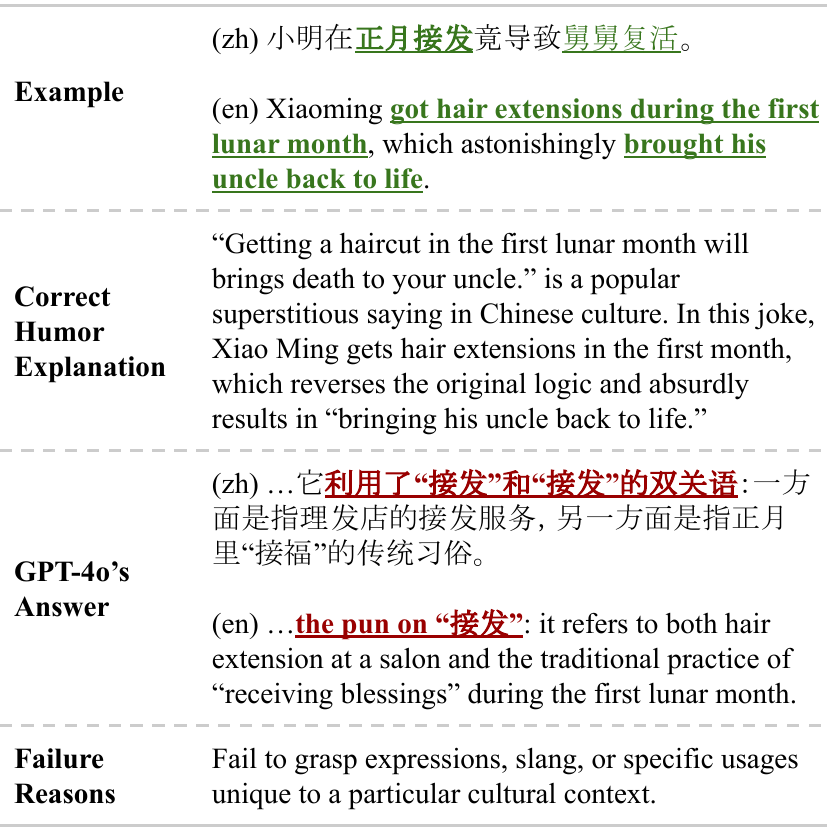}
    \caption{Culture unawareness example.}
    \label{fig:culture-example}
    \vspace{-0.5em}
\end{figure}

\begin{figure}[t]
    \includegraphics[width=\linewidth]{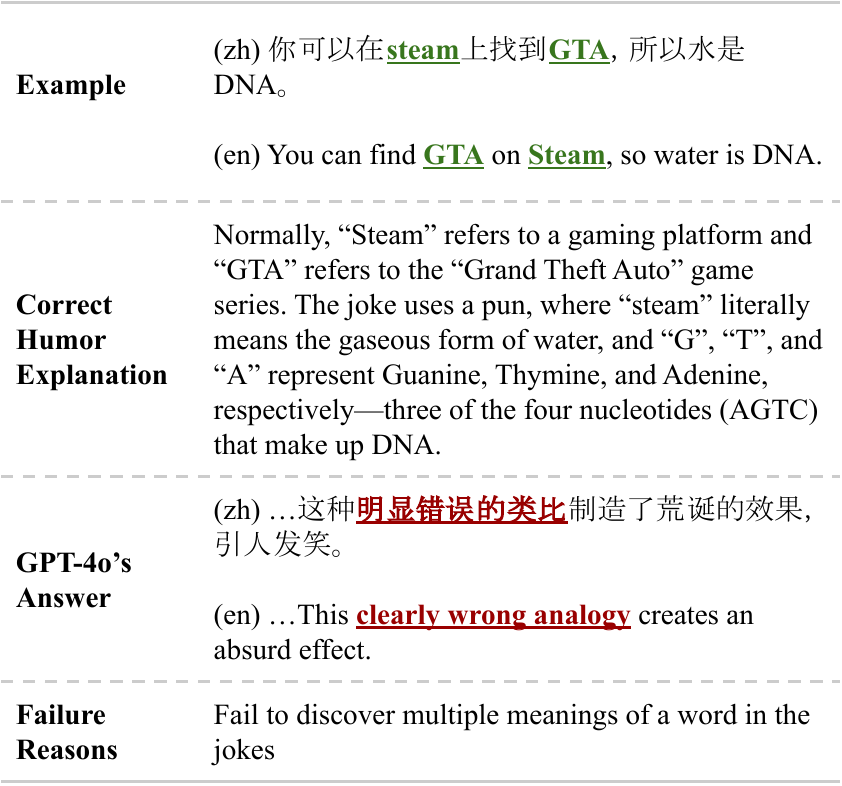}
    \caption{Pun-based error example.}
    \label{fig:pun-example}
    \vspace{-0.5em}
\end{figure}

\begin{figure}[t]
    \includegraphics[width=\linewidth]{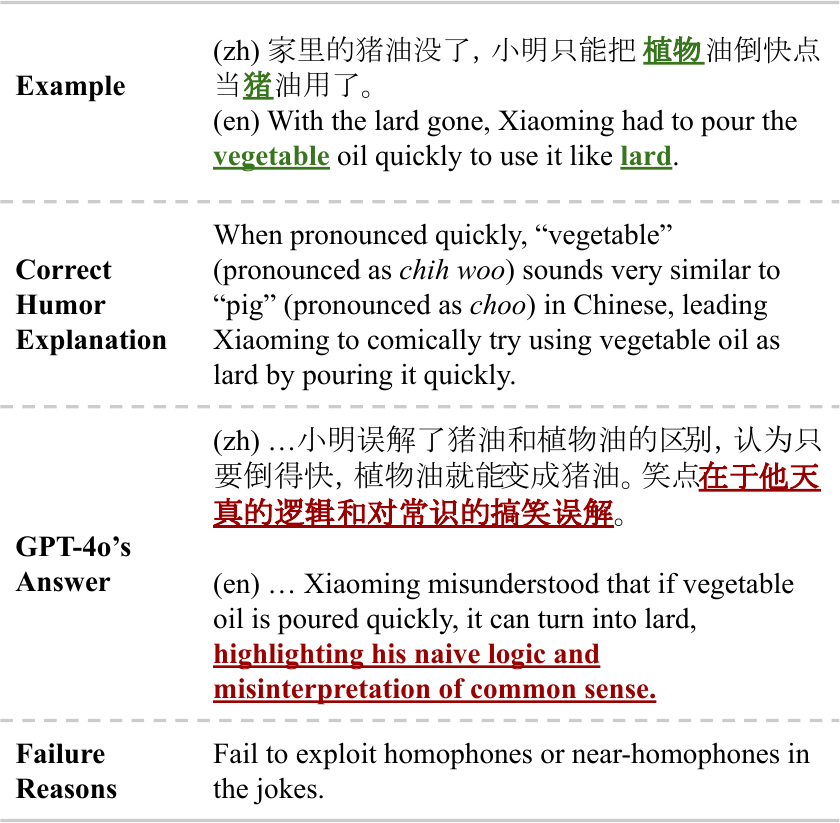}
    \caption{Homophonic error example.}
    \label{fig:homophonic-example}
    \vspace{-0.5em}
\end{figure}

\Cref{fig:distribution} shows the overall distribution of error types for GPT-4o and ERNIE\textsubscript{4-turbo} on \dataname~in terms of their humor explanations.
This error analysis is conducted by an individual who is not involved in writing the original explanations, ensuring an unbiased evaluation. 
GPT-4o is more prone to errors categorized as ``cultural unawareness'' (29.5\% of all its explanations) compared to ERNIE\textsubscript{4-turbo} (10.5\%).
We suspect that ERNIE\textsubscript{4-turbo} is more familiar with Chinese culture as it is likely trained on a larger Chinese corpus than GPT-4o.
However, GPT-4o performs better on cases requiring an understanding of contexts or puns, suggesting its strong reasoning ability.
We provide three error cases for GPT-4o here and additional cases for both GPT-4o and ERNIE\textsubscript{4-turbo} in \Cref{app-sec:more-error-cases}. 
In the following examples in \Cref{fig:culture-example}, \Cref{fig:pun-example} and \Cref{fig:homophonic-example}, we highlight key phrases that induce humor in green, and underscore the errors in red. 

\paragraph{Error Type I: Cultural Unawareness.}\leavevmode
LLMs may fail to explain a joke due to a lack of awareness of certain cultural knowledge.
For instance, the example in \Cref{fig:culture-example} requires knowledge of a superstitious belief in Chinese culture, {\it getting a haircut in the first lunar month brings death to your uncle}, and the explanation from GPT-4o fails to connect to this Chinese cultural belief.
We hypothesize that while LLMs are pre-trained on Internet-scale corpora, such culturally specific knowledge can still be challenging for them to grasp.
Moreover, even when they have acquired such cultural knowledge, they may fail to relate to it as we humans do during the reasoning process.

\paragraph{Error Type II: Pun-based Error.}

\noindent LLMs may fail to identify multiple meanings of a single word within a joke, causing them to fail on pun-based jokes where humor lies in inverting the conventional usage of words. 
In \Cref{fig:pun-example}, GPT-4o fails to grasp the transition from the video game terms ``Steam'', ``GTA'' to the scientific terminologies ``G'', ``T'', ``A'' that form DNA. 
Typically, ``Steam'' refers to a game platform, and ``GTA'' refers to the game series ``Grand Theft Auto''. 
The joke employs a pun on words where ``steam'' in its literal sense means water vapor, and ``GTA'' can represent not only the video game, but guanine, thymine, and adenine, which are nucleotides involved in the structure of DNA. 
Such jokes require LLMs to identify puns and the reason for the association of the multiple meanings.
Furthermore, the process requires LLMs to bridge the logic gap between these terms, such as ``steam'' and ``GTA'', and an unusual context, like ``water is DNA''. 
The overall process requires both scientific knowledge and creative thinking, which are challenging for LLMs.

\paragraph{Error Type III: Homophonic Error.}

The example in \Cref{fig:homophonic-example} requires LLMs to reason over the pronunciation as ``\chenv{植物}'' (pronounced as {\it chih woo}, meaning ``vegetable'') sounds very similar to ``\chenv{猪}'' (pronounced as {\it choo}, meaning ``pig'') in Chinese when we speak it fast enough. The humor arises from the contrast between the similarity in pronunciation and the disparity in meaning between the two terms.
Such contrasts may be sparse in the training corpus of LLMs, and also demand a deep connection across different modalities to link pronunciation with the meaning behind these terms, which poses significant challenges to LLMs.



\section{Conclusion}

We introduce \dataname, a Chinese humor understanding dataset that includes intellectually challenging and culturally specific humor in Chinese.
We show that \dataname~is challenging even for advanced LLMs and provide analysis of their failure cases.
We hope that \dataname~can advance non-English humor research and contribute to evaluating LLMs' reasoning abilities across diverse cultural backgrounds.

\section*{Limitations}
We try our best to test the Chinese humor understanding ability of different LLMs.
However, due to the limited budget and API access, we cannot evaluate all possible LLMs in this paper.
We encourage future research to conduct further evaluations of humor understanding abilities in LLMs.
In the meantime, we emphasize that our research focuses primarily on demonstrating how humor understanding remains a significant challenge, even for SOTA LLMs.
Our work shows that along with many other problems \cite{ignat-etal-2024-solved-open}, humor understanding, especially non-English and culturally specific humor understanding, remains an unsolved problem in the era of LLMs. 
We hope \dataname~can contribute to non-English humor understanding evaluations for future multilingual LLMs.

\section*{Ethics Statement}
We have made every effort to filter out excessively offensive content in RZB.
However, due to the subjective nature of humor, some of our jokes may still be perceived as offensive by individuals with different cultural or personal standards.
To address these concerns, we strongly recommend that researchers use \dataname~with cultural sensitivity, recognizing that the jokes in the dataset reflect the sociocultural context in which they were created.
We encourage users of \dataname~to approach the dataset with caution, remaining mindful of its potential to cause offense or harm, particularly when applying it in research or applications that involve diverse audiences or address sensitive topics.
We wish to foster an ethical and responsible approach to data collection and usage, and we welcome constructive feedback from the research community and stakeholders to continually improve \dataname~and mitigate potential harm.

\section{Acknowledgement}
The GPT experiments are supported by credit from OpenAI through OpenAI Researcher Access assigned to Naihao Deng. 
We appreciate Qiang Liu, and Xiaoyue Shi for helping with the human study.


\bibliography{iclr2024_conference,custom, aclanthology}

\clearpage

\appendix

\section{Contributions} 
\paragraph{Idea Proposal.}
Naihao Deng proposed the high-level idea of constructing a humor understanding benchmark sourced from RZB data.

\paragraph{Background Survey.} 
Ruiqi He surveyed the humor-related tasks. 

\paragraph{Data Processing.}
Ruiqi He crawled and processed the jokes from RZB.

\paragraph{Annotation.} 
Ruiqi He annotated the explanations for the RZB jokes.
Yushu He, Longju Bai, Jiarui Liu, Zhenjie Sun, Zhenghao Tang, He Wang, Naihao Deng conducted the preference annotations.

\paragraph{Experiments.}
Ruiqi He, Hanchen Xia, and Naihao Deng conducted the experiments.

\paragraph{Result Aggregation.}
Ruiqi He, Naihao Deng, Yushu He aggregated the results.

\paragraph{Paper Writing.}
Ruiqi He and Naihao Deng drafted the paper.
Other authors provided revisions and feedback on the paper.

\noindent Naihao Deng organized the research.

\section{Agreement Rate Calculation}
\label{app-sec: agreement-rate}
We calculate the percentage agreement rate among annotators who annotate their preferences between explanations from LLMs and humans.
The results show an average inter-annotator agreement of 61.9\% for GPT-4o and 60.9\% for ERNIE\textsubscript{4-turbo}. 
Given the inherent subjectivity of humor interpretation tasks \cite{deng-etal-2023-annotate}, the combined average agreement percentage of 61.4\% is decent.

\section{Annotation Instructions for Preference Annotation}
\label{app-sec: annotation-instruction}

\noindent We include the following instructions for the preference annotations of the joke explanations:








\begin{mybox}{Instruction}
\chenv{\noindent ``在这个标注中，你将会看到一个笑话和对这个笑话的幽默之处的两个解释，请你比较哪个解释更好的解释了这个笑话的幽默之处，并从以下三个标签中选择：

\noindent 1. 解释1 

\noindent 2. 解释2 

\noindent 3. 一样好''}
\end{mybox}

\noindent which translates to,

\begin{mybox}{Instruction Translation}
\chenv{\noindent ``In this annotation task, you will see a joke along with two explanations of its humor. 
Please compare which explanation better explains the reason why this joke is funny and choose from the following three labels:

\noindent 1. Explanation 1

\noindent 2. Explanation 2

\noindent 3. Both are equally good.''}
\end{mybox}

For each example, we randomly assign the explanations from the LLMs and the human as Explanation 1 and Explanation 2 to ensure a fair comparison.

\section{More Error Cases}
\label{app-sec:more-error-cases}

We note that many examples here encompass multiple error types, highlighting the complexity of \dataname.

\paragraph{Insufficient Contextual Understanding.}\leavevmode
%
%
\noindent
\begin{figure}[!ht]
    \centering
    \includegraphics[width=\linewidth]{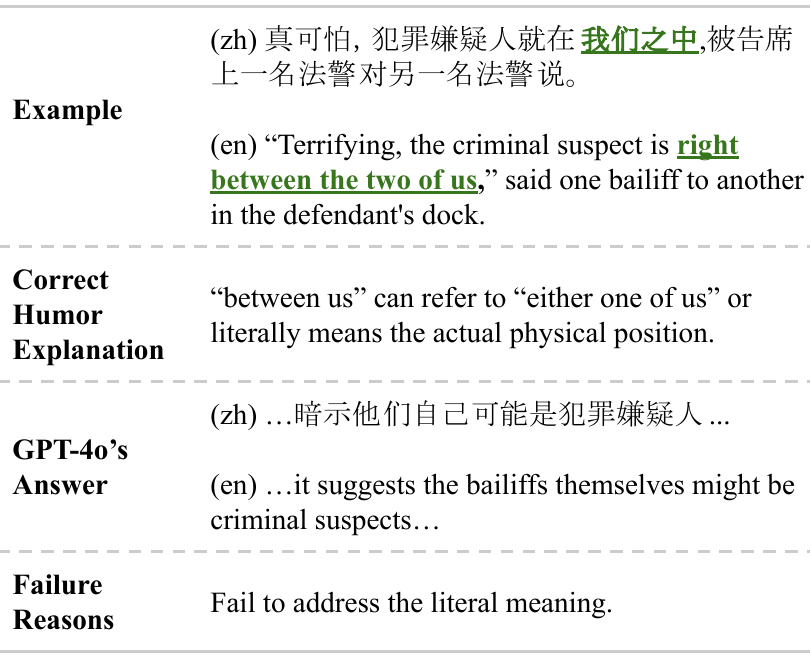}
    \caption{Insufficient contextual understanding example.}
    \label{fig:situational-example}
\end{figure}
\noindent LLMs may fail to ground their responses to the context when they explain the joke.
For instance, in the example in \Cref{fig:situational-example}, ``between us'' typically means ``either you or me'', but it also has the literal meaning to indicate the person standing ``between us'', which is the right interpretation given that the two bailiffs are talking about the criminal.
However, GPT-4o only reasons that ``the criminal is either you or me'' but fails to capture the literal meaning from the context.
We hypothesize that in the pre-training corpus, ``between us'' most likely acquires the meaning of ``either you or me'' rather than the literal meaning in a scenario like this, which creates a bias that prevents the model from reasoning about the literal interpretation required for this specific explanation.

\begin{figure}[!ht]
    \centering
    \includegraphics[width=\linewidth]{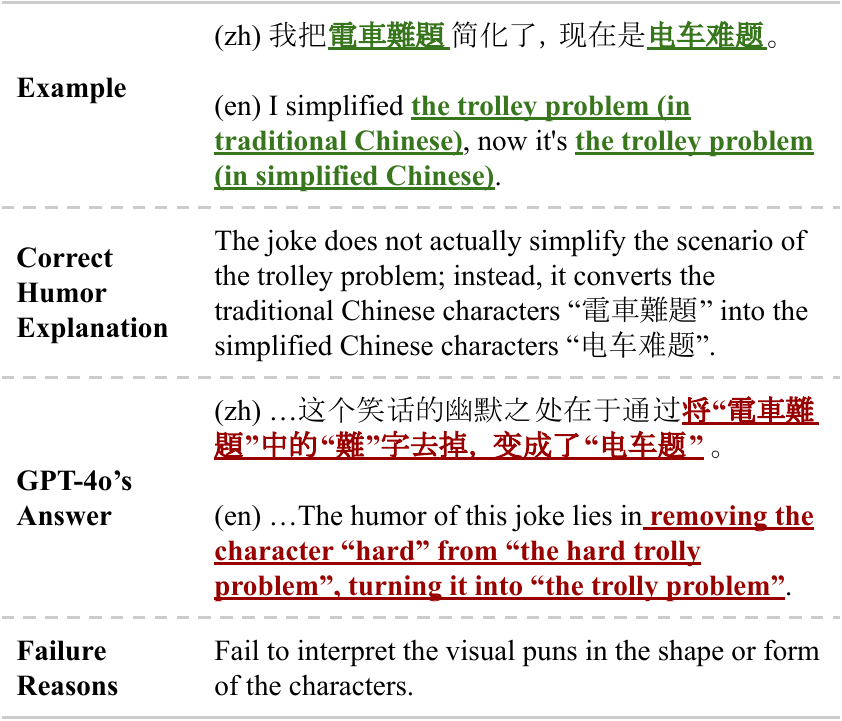}
    \caption{Glyph-based error example.}
    \label{fig:glyph-example}
\end{figure}

\paragraph{Glyph-based Error.}\leavevmode
\noindent LLMs may fail to interpret the visual puns based on the shape or form of Chinese characters. Glyph-based humor in Chinese leverages its logographic writing system, where characters integrate both semantic and visual elements. Unlike the phonemic alphabet used in English, Chinese characters’ pictorial and ideographic nature allows for visual puns in jokes \citep{daniels1996world}.
In the example in \Cref{fig:glyph-example}, ``simplify'' does not refer to simplifying the trolly problem conceptually, but to simplifying the traditional Chinese characters to simplified Chinese characters as the traditional Chinese characters are also termed as ``complicated characters''.
However, LLMs struggle to reason such graphemic differences as there are no explicit connections between the textual meaning and visual representations of the glyphs.

\begin{figure}[!ht]
    \centering
    \includegraphics[width=\linewidth]{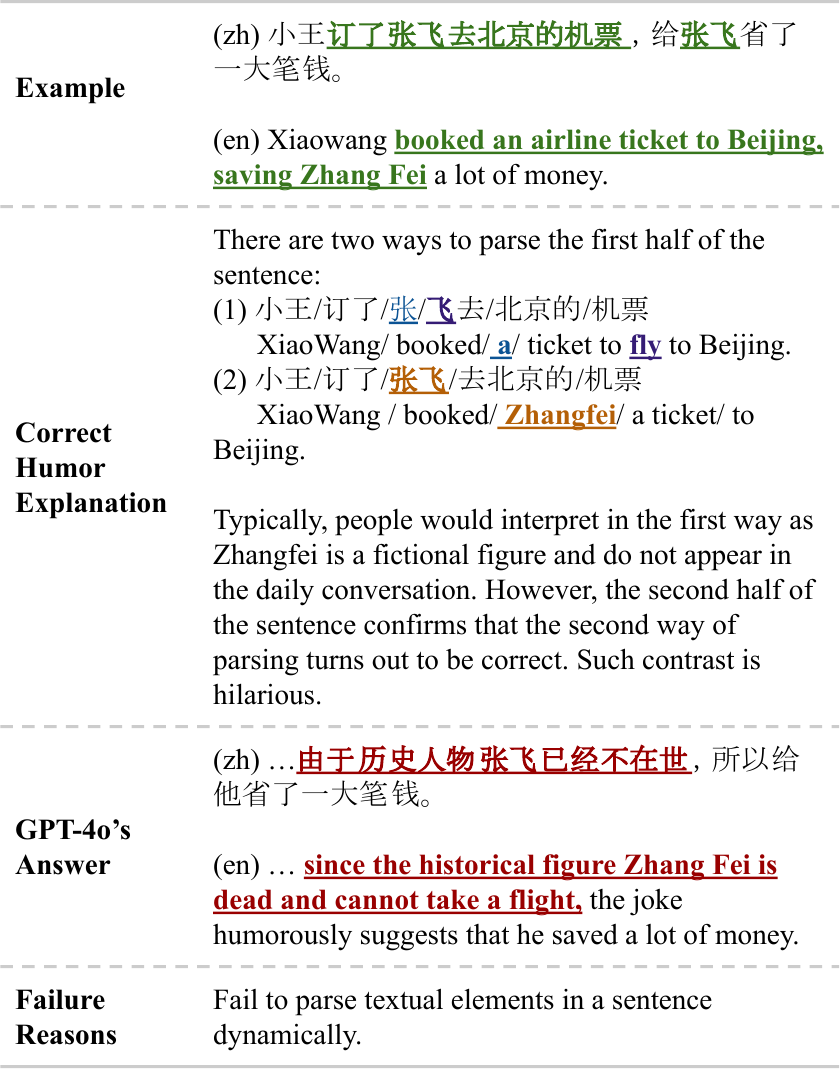}
    \caption{Parsing error example.}
    \label{fig:parsing-error-example}
\end{figure}

\paragraph{Parsing Error.}\leavevmode
\noindent LLMs often fail to parse sentences in multiple ways simultaneously, leading to difficulties in explaining jokes that require different parsing for the same sentence. 
In the example in \Cref{fig:parsing-error-example}, the humor hinges on the ambiguity of the phrase ``\chenv{张飞}'', which can be interpreted either as part of a verb phrase implying ``a ticket flying to Beijing'' or as a proper noun, referring to the historical figure Zhang Fei. 
This ambiguity stems from the flexibility of the Chinese language, where each character can function independently as a word or combine with others to form new words or phrases. 
There are decades of research studying the problem of parsing Chinese \cite{sun-jurafsky-2004-shallow, sun-etal-2009-chinese}.
Recently, researchers have proposed task-specific tokenization approaches that adapt the parsing process to better align with downstream tasks \cite{liu-etal-2021-bridging, liu-etal-2023-task}.
However, how to incorporate different ways of parsing at one time still remains challenging.

\begin{figure}[!ht]
    \centering
    \includegraphics[width=\linewidth]{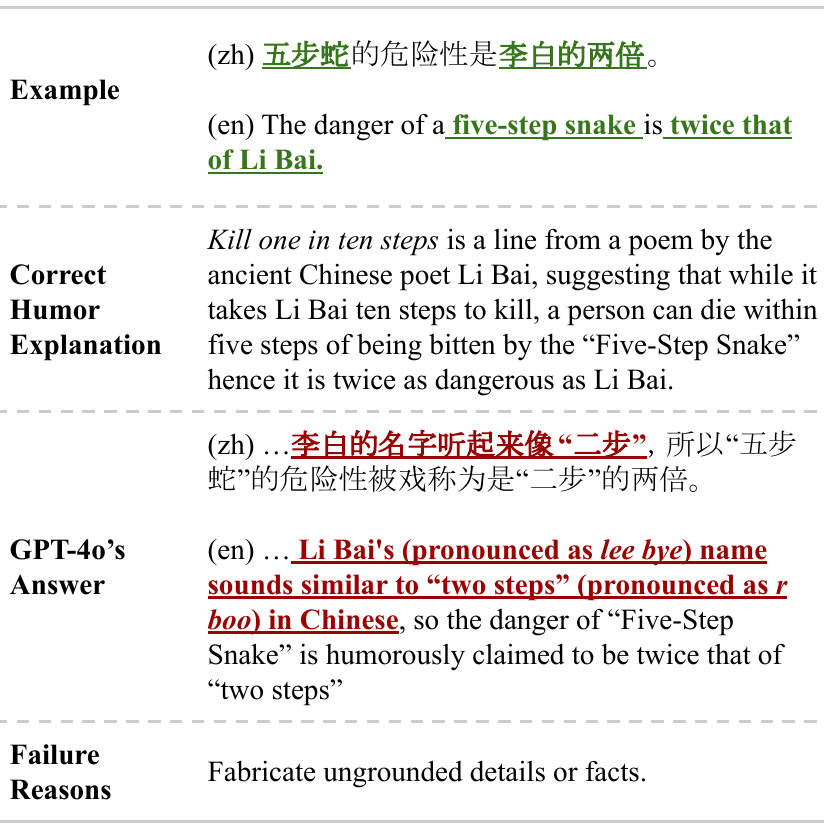}
    \caption{Hallucinations example.}
    \label{fig:hallu-example}
\end{figure}

\paragraph{Hallucinations.}\leavevmode
\noindent LLMs may fabricate ungrounded details or facts in joke explanations. 
For instance, in the explanation in \Cref{fig:hallu-example}, GPT-4o claims that  ``Li Bai's name sounds similar to two steps'', while ``Li Bai'' (pronounced as {\it lee bye}) does not sound like ``two steps'' (pronounced as {\it r boo}).

On the other hand, the correct explanation requires an understanding of a Chinese poem from Li Bai, ``\chenv{十步杀一人}'' (The warrior kills a person for every ten steps). 
This line praises the courage of the soldiers, but the joke deliberately portrays this as a characteristic of Li Bai.
Therefore, compared to Li Bai who can kill a person in ten steps, a five-step snake, which can kill a person in five steps, is twice as dangerous as Li Bai.
Such explanation requires LLMs to have a deep understanding of Chinese culture and reason over  cultural references, posing a great challenge to current LLMs.
Although recent works have made progress towards building LLMs beyond English \cite{du2024chinese, zhao2024llama}, building an LLM that can comprehend such nuanced Chinese cultural terms can be extremely hard.
\newpage

\begin{figure}[!ht]
    \centering
    \includegraphics[width=\linewidth]{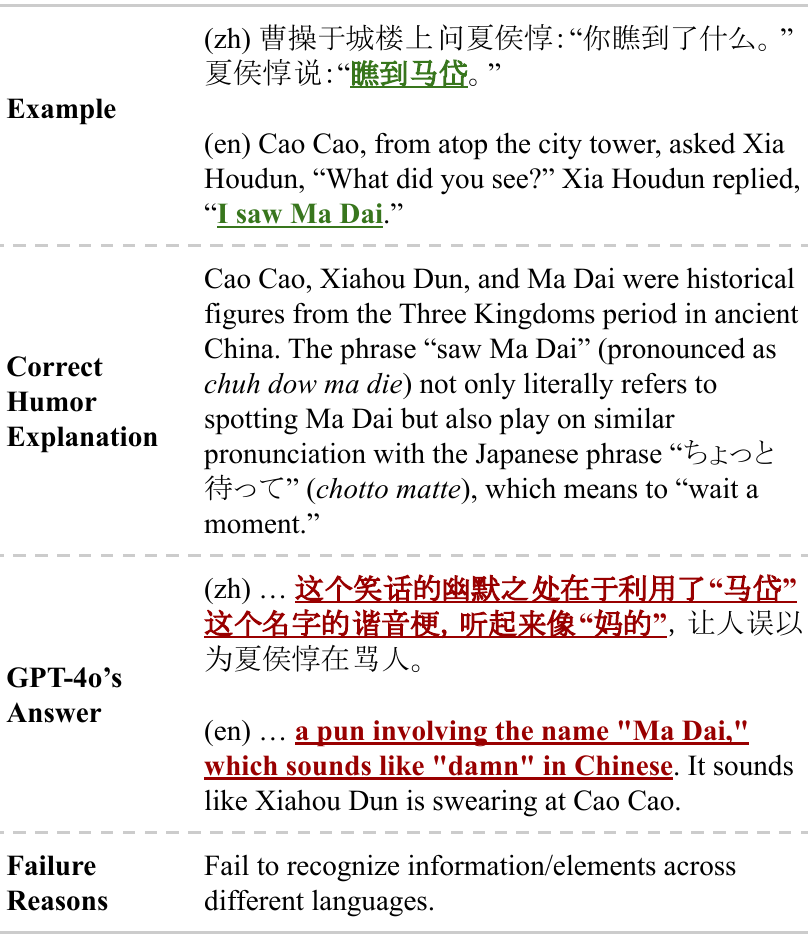}
    \caption{Cross-lingual error example.}
    \label{fig:cross-example}
\end{figure}

\paragraph{Cross-lingual Error.}\leavevmode
\noindent LLMs may fail to recognize elements or information across different languages.
In the explanation in \Cref{fig:cross-example}, GPT-4o attempts to link the pronunciation of ``Ma Dai'' to other Chinese terms but fails to identify the similar pronunciations across the Chinese term ``\chenv{瞧到马岱}'' (pronounced as {\it chuh dow ma die}, meaning ``saw Ma Dai'')  and the Japanese term ``\raisebox{-0.15cm}{\includegraphics[width=0.3\linewidth]{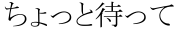}}'' ({\it chotto matte}, meaning ``wait a moment'').
Such cases require LLMs to connect pronunciations across languages, which may be rare in the LLMs' pre-training corpus and poses significant challenges to current LLMs.


\paragraph{Overcritical.}

\begin{figure}[!ht]
    \includegraphics[width=\linewidth]{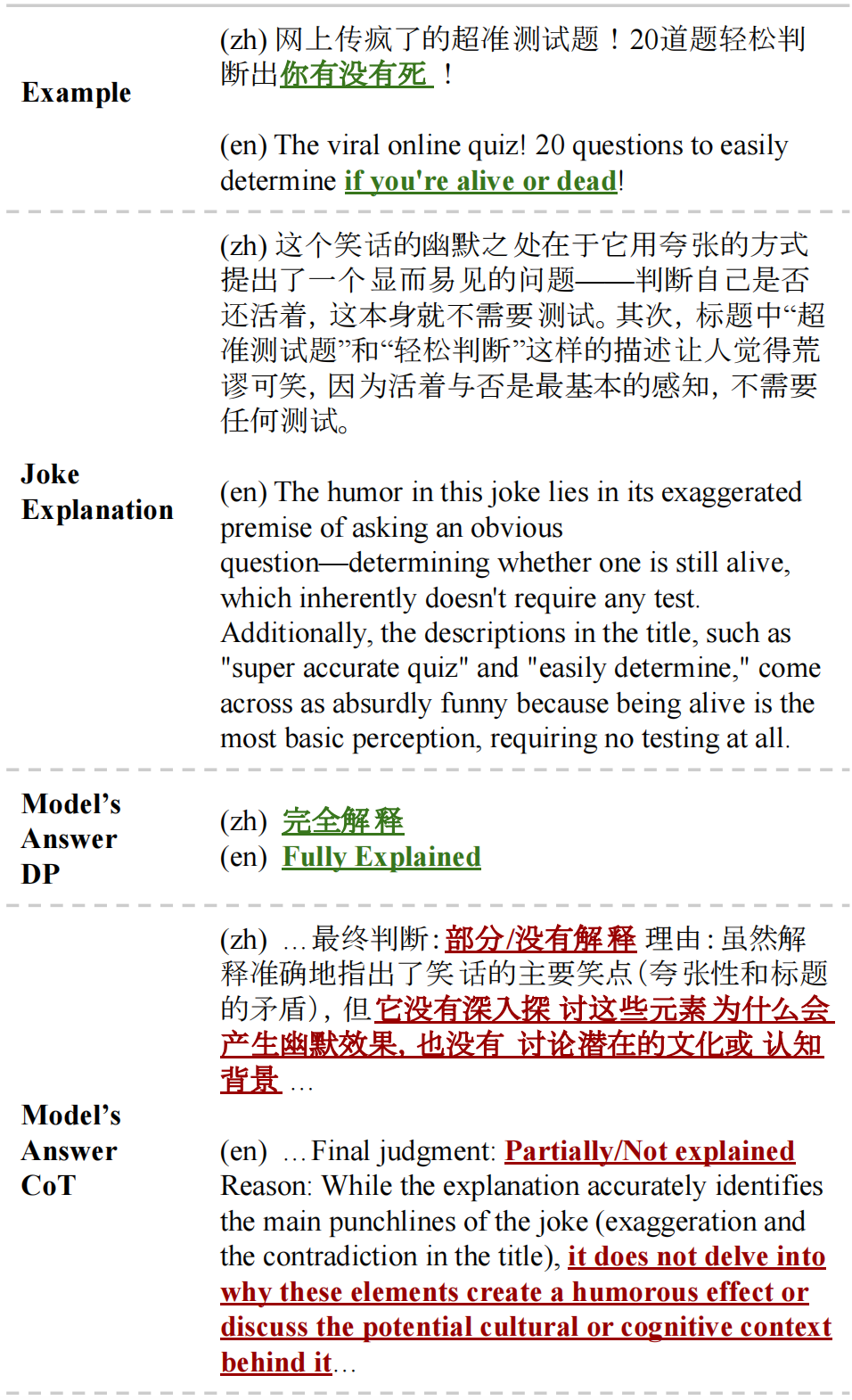}
    \caption{Overcritical example by Nemotron\textsubscript{70B}. The Nemotron\textsubscript{70B} model selects the correct answer in the DP prompting, but selects the incorrect answer due to being overly critical in the CoT prompting.}
    \label{fig:over_critical_ex}
\end{figure}

\Cref{fig:over_critical_ex} shows when the model overly criticizes an explanation, it focuses excessively on minor details, neglecting the major point and ultimately misjudging the explanation. The joke relies on common sense about life, death, and testing rather than cultural knowledge. Under the CoT prompting, the model recognizes that the explanation identifies the main punchlines but overly criticizes the lack of cultural context.

\paragraph{Excessive Sensitivity.} 

For ERNIE\textsubscript{4-turbo},  in addition to errors across all the aforementioned categories, it also demonstrates excessive sensitivity to certain examples.
Specifically, when  content includes languages related to hate speech but used in non-harmful contexts, ERNIE\textsubscript{4-turbo} refuses to provide an explanation.
During our evaluation, we observe this excessive sensitivity in the ERNIE\textsubscript{4-turbo}'s responses to humor related to medical ethics and political discussions.
This suggests that correctly understanding the context and the language toxicity remains an open challenge \cite{zhang2024dont}.
Such issues are particularly critical for humor explanation, as misclassifying non-toxic context can cause the responses to deviate from the intended humor.

\section{Prompts for DP and CoT in \texorpdfstring{\dataname}{Chumor CLS}}
\label{app-sec: prompt}

This section outlines the prompts used in \dataname\ to evaluate whether an explanation fully explains a joke. Two prompting strategies are adopted: Direct Prompting (DP) and Chain of Thought (CoT). Below are the details of  each approach:

\begin{mybox}{Direct Prompting (DP)}
\chenv{
    你将看到一个笑话以及对这个笑话的解释。请判断这个解释是否完全解释了笑话。
    根据判断，选择“完全解释”或“部分/没有解释”，不需要解释为什么对或者不对。

    笑话：\{joke\} \\
    笑话解释：\{explanation\}
}
\end{mybox}

\begin{mybox}{Translations}
You will see a joke and an explanation of the joke. Please determine whether this explanation fully explains the joke. Based on your judgment, choose either "fully explains" or "partially/does not explain." You do not need to explain why it is correct or incorrect.

Joke: \{joke\} \\
Explanation: \{explanation\}
\end{mybox}

The DP prompt is designed to encourage concise decision-making. It directly asks the model to evaluate the completeness of the explanation without requiring reasoning or justification.

\begin{mybox}{Chain of Thought (CoT)}
\chenv{
    你将看到一个笑话以及对这个笑话的解释。请逐步思考，写下过程并最终判断这个解释是否完全解释了笑话。
    根据判断，选择“完全解释”或“部分/没有解释”。

    笑话：\{joke\} \\
    笑话解释：\{explanation\}
}
\end{mybox}

\begin{mybox}{Translations}
You will see a joke and an explanation of the joke. Please think step by step, write down your reasoning process, and finally determine whether this explanation fully explains the joke. Based on your judgment, choose either "fully explains" or "partially/does not explain."

Joke: \{joke\} \\
Explanation: \{explanation\}"
\end{mybox}

The CoT prompt, in contrast, requires the model to reason step by step before reaching a conclusion. This approach aims to improve transparency by explicitly documenting the thought process behind the evaluation.

\section{Detailed Results of Experiments}
\label{app-sec: detailed results}
For evaluation, we input each prompt into the model and collect its responses, comparing them to the labels in \dataname. A model's response is considered correct if it matches the reference label. If the model provides an incorrect answer or doesn't generate a response at all (due to safety protocols or filtering sensitive terms), it is marked as incorrect. Such scenario is rare, occurring only 21 times in our experiments, and exclusively with GLM-4\textsubscript{plus}.

\begin{figure}[!ht]
    \includegraphics[width=\linewidth]{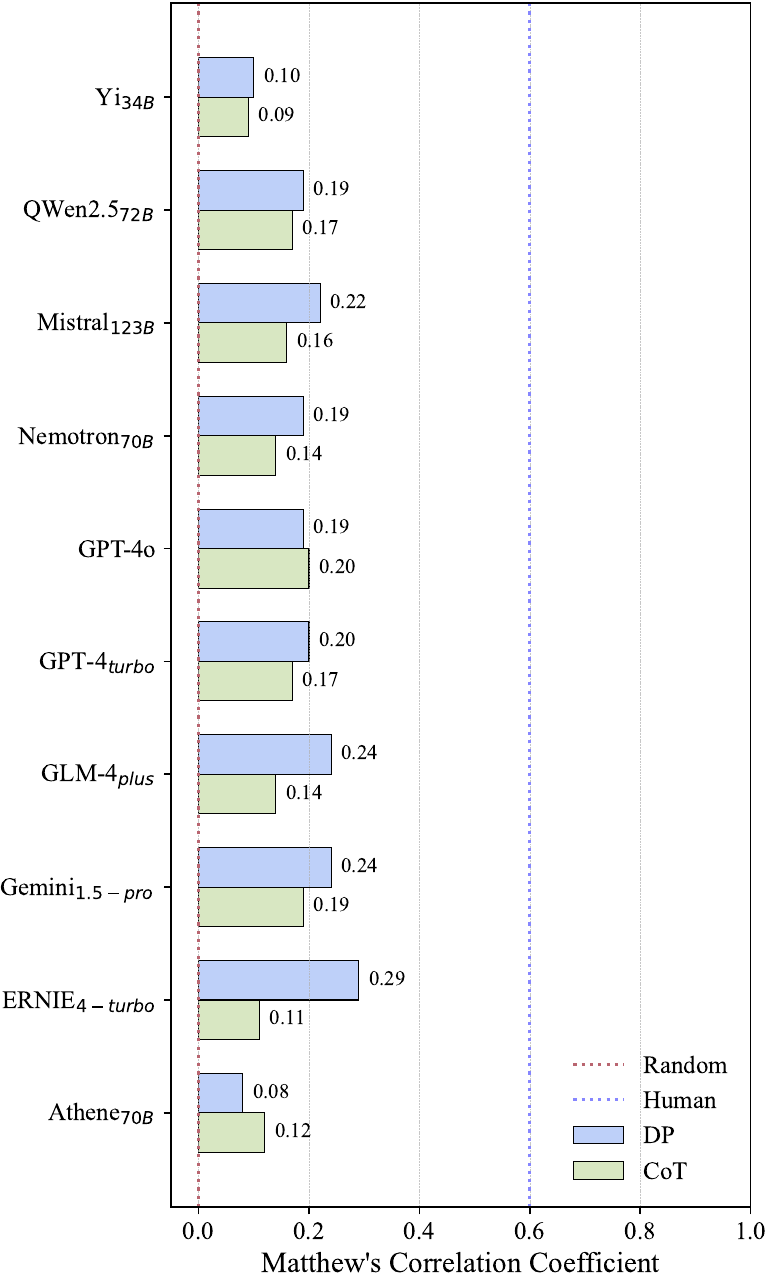}
    \caption{The Matthew's correlation coefficient of different models' test results in DP and CoT.}
    \label{fig: MCC of different models}
\end{figure}

We highlight that CoT prompting at most cases degrade the models' performance on \dataname. As shown in \Cref{fig: MCC of different models}, only Athene\textsubscript{70B} achieves a significant improvement. 
However, this is offset by its poorest performance under DP prompting among the models. 
GPT-4o shows a slight improvement, with its MCC score increasing from 0.19 to 0.20. And all other eight models exhibit different degrees of performance decline.


\begin{figure}[!ht]
    \includegraphics[width=\linewidth]{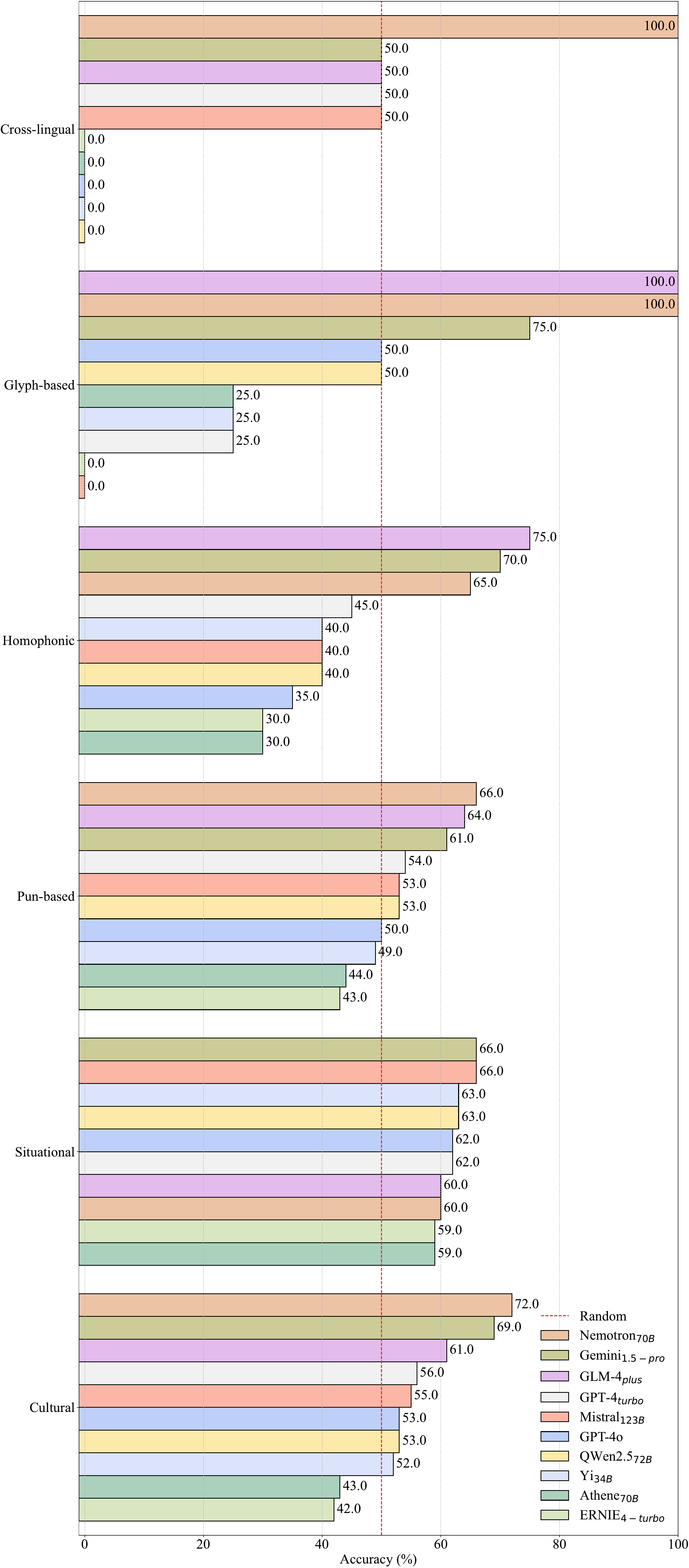}
    \caption{CoT accuracy on different joke types(\%).}
    \label{fig:cot}
\end{figure}

\begin{table*}[!ht]
    \centering
    \small
    \renewcommand{\arraystretch}{1.3}
    \setlength\tabcolsep{3pt}
    \begin{tabular}{@{}lcccccccc@{}}
        \toprule
        \multirow{2}{*}{Model} & \multicolumn{4}{c}{DP} & \multicolumn{4}{c}{CoT} \\ \cmidrule(lr){2-5} \cmidrule(lr){6-9}
                              & MCC & ACC (\%) & FPR (\%) & FNR (\%) & MCC & ACC (\%) & FPR (\%) & FNR (\%) \\ \midrule
        Yi\textsubscript{34B} & 0.10 & 44.95 & 97.24 & 0.21 & 0.09 & 47.17 & 89.30 & 5.44 \\ 
        Nemotron\textsubscript{70B} & 0.19 & 56.30 & 61.26 & 20.87 & 0.14 & 57.17 & 40.28 & 46.14 \\
        Athene\textsubscript{70B} & 0.08 & 44.59 & 97.83 & 0.28 & 0.12 & 47.26 & 91.10 & 2.89 \\
        ERNIE\textsubscript{4-turbo} & 0.29 & 60.29 & 59.83 & 13.57 & 0.11 & 45.16 & 96.93 & 0.14 \\
        QWen2.5\textsubscript{72B} & 0.19 & 48.46 & 90.67 & 0.69  & 0.17 & 49.45 & 86.91 & 3.31 \\
        Mistral\textsubscript{123B} & 0.22 & 55.56 & 69.26 & 12.19 & 0.16 & 51.18 & 79.92 & 8.40 \\
        Gemini\textsubscript{1.5-pro} & 0.24 & 54.00 & 77.42 & 5.17  & 0.19 & 60.32 & 33.81 & 47.31 \\
        GLM-4\textsubscript{plus} & 0.24 & 55.56 & 72.28 & 8.26 & 0.14 & 58.13 & 32.96 & 53.44 \\
        GPT-4o & 0.19 & 51.87 & 80.02 & 6.68 & 0.20 & 50.64 & 85.00 & 3.03 \\
        GPT-4\textsubscript{turbo} & 0.20 & 52.32 & 79.28 & 6.61 & 0.17 & 51.27 & 80.87 & 6.96 \\
        \bottomrule
    \end{tabular}
    \caption{Performance metrics for explanation evaluation including Matthew's correlation coefficient (MCC), accuracy (ACC), false positive rate (FPR), and false negative rate (FNR).}
    \label{table:performance_categorization}
\end{table*}

\begin{table*}[t]
    \centering
    \small
    \begin{tabular}{@{}llcccccccc@{}}
        \toprule
        \multirow{2}{*}{Model} & \multirow{2}{*}{Source} & \multicolumn{4}{c}{DP} & \multicolumn{4}{c}{CoT} \\ \cmidrule(lr){3-6} \cmidrule(lr){7-10}
                               &                         & MCC & ACC(\%) & FPR(\%) & FNR(\%) & MCC & ACC(\%) & FPR(\%) & FNR(\%) \\ \midrule
        \multirow{3}{*}{Athene\textsubscript{70B}} 
           & Overall   & 0.08 & 44.59 & 97.83 & 0.28 & 0.12 & 47.26 & 91.10 & 2.89 \\
           & ERNIE Bot & 0.12 & 52.38 & 97.15 & 0.00 & 0.15 & 54.24 & 91.13 & 2.13 \\
           & GPT-4o    & 0.03 & 33.90 & 98.51 & 0.86 & 0.08 & 37.67 & 91.06 & 4.50 \\ \midrule

        \multirow{3}{*}{ERNIE\textsubscript{-turbo}} 
           & Overall   & 0.29 & 60.29 & 59.83 & 13.57 & 0.11 & 45.16 & 96.93 & 0.14 \\
           & ERNIE Bot & 0.23 & 58.64 & 78.14 & 5.99  & 0.16 & 53.47 & 94.83 & 0.10 \\
           & GPT-4o    & 0.27 & 62.54 & 41.38 & 29.55 & 0.04 & 33.76 & 99.04 & 0.21 \\ \midrule

        \multirow{3}{*}{Gemini\textsubscript{1.5-pro}} 
           & Overall   & 0.24 & 54.00 & 77.42 & 5.17 & 0.19 & 60.32 & 33.81 & 47.31 \\
           & ERNIE Bot & 0.27 & 60.66 & 74.13 & 5.89 & 0.23 & 60.87 & 28.62 & 49.24 \\
           & GPT-4o    & 0.21 & 44.85 & 80.74 & 3.64 & 0.17 & 59.56 & 39.04 & 43.25 \\ \midrule

        \multirow{3}{*}{GLM-4\textsubscript{plus}}
           & Overall   & 0.24 & 55.56 & 72.28 & 8.26 & 0.14 & 58.13 & 32.96 & 53.44 \\
           & ERNIE Bot & 0.25 & 59.83 & 74.97 & 6.70 & 0.15 & 57.56 & 37.06 & 47.61 \\
           & GPT-4o    & 0.21 & 49.68 & 69.57 & 11.56 & 0.06 & 58.92 & 28.83 & 65.74 \\ \midrule

        \multirow{3}{*}{GPT-4\textsubscript{turbo}}
           & Overall   & 0.20 & 52.32 & 79.28 & 6.61 & 0.17 & 51.27 & 80.87 & 6.96 \\
           & ERNIE Bot & 0.20 & 57.25 & 80.99 & 5.99 & 0.22 & 58.75 & 76.14 & 7.72 \\
           & GPT-4o    & 0.18 & 45.56 & 77.55 & 7.92 & 0.13 & 41.01 & 85.64 & 5.35 \\ \midrule

        \multirow{3}{*}{GPT-4o}
           & Overall   & 0.19 & 51.87 & 80.02 & 6.68 & 0.20 & 50.64 & 85.00 & 3.03 \\
           & ERNIE Bot & 0.21 & 57.82 & 79.41 & 6.40 & 0.24 & 58.07 & 82.47 & 2.94 \\
           & GPT-4o    & 0.16 & 43.71 & 80.64 & 7.28 & 0.15 & 40.44 & 87.55 & 3.21 \\ \midrule

        \multirow{3}{*}{Nemotron\textsubscript{70B}}
           & Overall   & 0.19 & 56.30 & 61.26 & 20.87 & 0.14 & 57.17 & 40.28 & 46.14 \\
           & ERNIE Bot & 0.22 & 60.66 & 56.81 & 22.54 & 0.14 & 57.04 & 39.18 & 46.60 \\
           & GPT-4o    & 0.18 & 50.32 & 65.74 & 17.34 & 0.13 & 57.36 & 41.38 & 45.18 \\ \midrule

        \multirow{3}{*}{Mistral\textsubscript{123B}}
           & Overall   & 0.22 & 55.56 & 69.26 & 12.19 & 0.16 & 51.18 & 79.92 & 8.40 \\
           & ERNIE Bot & 0.25 & 61.13 & 65.15 & 13.60 & 0.18 & 57.04 & 79.73 & 7.61 \\
           & GPT-4o    & 0.20 & 47.90 & 73.40 & 9.21 & 0.12 & 43.14 & 80.11 & 10.06 \\ \midrule

        \multirow{3}{*}{Qwen2.5\textsubscript{72B}}
           & Overall   & 0.19 & 48.46 & 90.67 & 0.69 & 0.17 & 49.45 & 86.91 & 3.31 \\
           & ERNIE Bot & 0.19 & 54.45 & 92.61 & 0.30 & 0.18 & 55.54 & 88.07 & 2.54 \\
           & GPT-4o    & 0.17 & 40.23 & 88.72 & 1.50 & 0.14 & 41.08 & 85.74 & 4.93 \\ \midrule

        \multirow{3}{*}{Yi\textsubscript{34B}}
           & Overall   & 0.10 & 44.95 & 97.24 & 0.21 & 0.09 & 47.17 & 89.30 & 5.44 \\
           & ERNIE Bot & 0.15 & 53.42 & 94.72 & 0.30 & 0.11 & 53.99 & 88.38 & 5.28 \\
           & GPT-4o    & 0.03 & 33.33 & 99.79 & 0.00 & 0.07 & 37.81 & 90.21 & 5.78 \\
        \bottomrule
    \end{tabular}
    \caption{Detailed performance metrics with source for explanation evaluation of Matthew's correlation coefficient (MCC), accuracy (ACC), false positive rate (FPR), and false negative rate (FNR).}
    \label{table:performance_metrics}
\end{table*}





\end{document}